\documentclass[letterpaper]{article} 
\usepackage{aaai24}  
\usepackage{times}  
\usepackage{helvet}  
\usepackage{courier}  
\usepackage[hyphens]{url}  
\usepackage{graphicx} 
\usepackage{natbib}  
\usepackage{caption} 
\usepackage{algorithm}
\usepackage{algorithmicx}
\usepackage[noend]{algpseudocode}
\usepackage{bbding}

\usepackage{color}
\usepackage{xcolor}
\usepackage{appendix}
\usepackage{subcaption}

\usepackage{bm}
\usepackage{amsmath}
\usepackage{amssymb}
\usepackage{amsthm}

\usepackage[capitalise]{cleveref}

\usepackage{multirow}
\usepackage{multicol}

\usepackage{adjustbox}
\usepackage{booktabs}
\usepackage{threeparttable}
\usepackage{tabularx}

\usepackage{pifont}
\usepackage{mathtools}
\usepackage{thmtools}
\usepackage{thm-restate}

\newcommand{\xB}{\mathbf{x}}
\newcommand{\yB}{\mathbf{y}}
\newcommand{\zB}{\mathbf{z}}

\newcommand{\Dcal}{\mathcal{D}}

\newcommand{\Fcal}{\mathcal{F}}
\newcommand{\Hcal}{\mathcal{H}}
\newcommand{\Lcal}{\mathcal{L}}
\newcommand{\Ncal}{\mathcal{N}}
\newcommand{\Ocal}{\mathcal{O}}

\newcommand{\Scal}{\mathcal{S}}

\newcommand{\Rbb}{\mathbb{R}}

\newcommand{\papertitle}{G-NAS: Generalizable Neural Architecture Search for Single Domain Generalization Object Detection}
\newcommand{\methodname}{G-NAS}

\newcommand{\wu}[1]{{\textcolor{black}{#1}}}

\newcommand{\ie}{i.e.}
\newcommand{\eg}{\textit{e.g.}}

\usepackage{newfloat}
\usepackage{listings}
\DeclareCaptionStyle{ruled}{labelfont=normalfont,labelsep=colon,strut=off} 
\floatstyle{ruled}
\newfloat{listing}{tb}{lst}{}
\floatname{listing}{Listing}
\title{\papertitle}
\author{
    Fan Wu\textsuperscript{\rm 1},
    Jinling Gao\textsuperscript{\rm 1},
    Lanqing Hong\textsuperscript{\rm 2},
    Xinbing Wang\textsuperscript{\rm 1},
    Chenghu Zhou\textsuperscript{\rm 1},
    Nanyang Ye\textsuperscript{\rm 1}\thanks{Nanyang Ye is the corresponding author.}
}
\affiliations{
    \textsuperscript{\rm 1} Shanghai Jiao Tong University, Shanghai, China\\
    \textsuperscript{\rm 2} Huawei Noah's Ark Lab, Hong Kong, China\\

    wufan55@sjtu.edu.cn, gaojinling@sjtu.edu.cn, honglanqing@huawei.com, xwang8@sjtu.edu.cn, zhouchsjtu@gmail.com, ynylincoln@sjtu.edu.cn
}

\begin{document}

\maketitle

\begin{abstract}
In this paper, we focus on a realistic yet challenging task, Single Domain Generalization Object Detection (S-DGOD), where only one source domain's data can be used for training object detectors, but have to generalize multiple distinct target domains.
In S-DGOD, both high-capacity fitting and generalization abilities are needed due to the task's complexity. 
Differentiable Neural Architecture Search (NAS) is known for its high capacity for complex data fitting \wu{and we propose to leverage Differentiable NAS to solve S-DGOD.}
\wu{However, it may confront severe over-fitting issues due to the feature imbalance phenomenon, where parameters optimized by gradient descent are biased to learn from the easy-to-learn features, which are usually non-causal and spuriously correlated to ground truth labels, such as the features of background in object detection data.
Consequently, this leads to serious performance degradation, especially in generalizing to unseen target domains with huge domain gaps between the source domain and target domains. 
To address this issue, we propose the Generalizable loss (G-loss), which is an OoD-aware objective, preventing NAS from over-fitting by using gradient descent to optimize parameters not only on a subset of easy-to-learn features but also the remaining predictive features for generalization, and the overall framework is named~\methodname.
Experimental results on the S-DGOD urban-scene datasets demonstrate that the proposed~\methodname~achieves SOTA performance compared to baseline methods.
Codes are available at \textit{https://github.com/wufan-cse/G-NAS}.
}
\end{abstract}

\section{Introduction}
\label{sec:introduction}

\begin{figure}[t]
    \centering
    \includegraphics[width=\linewidth]{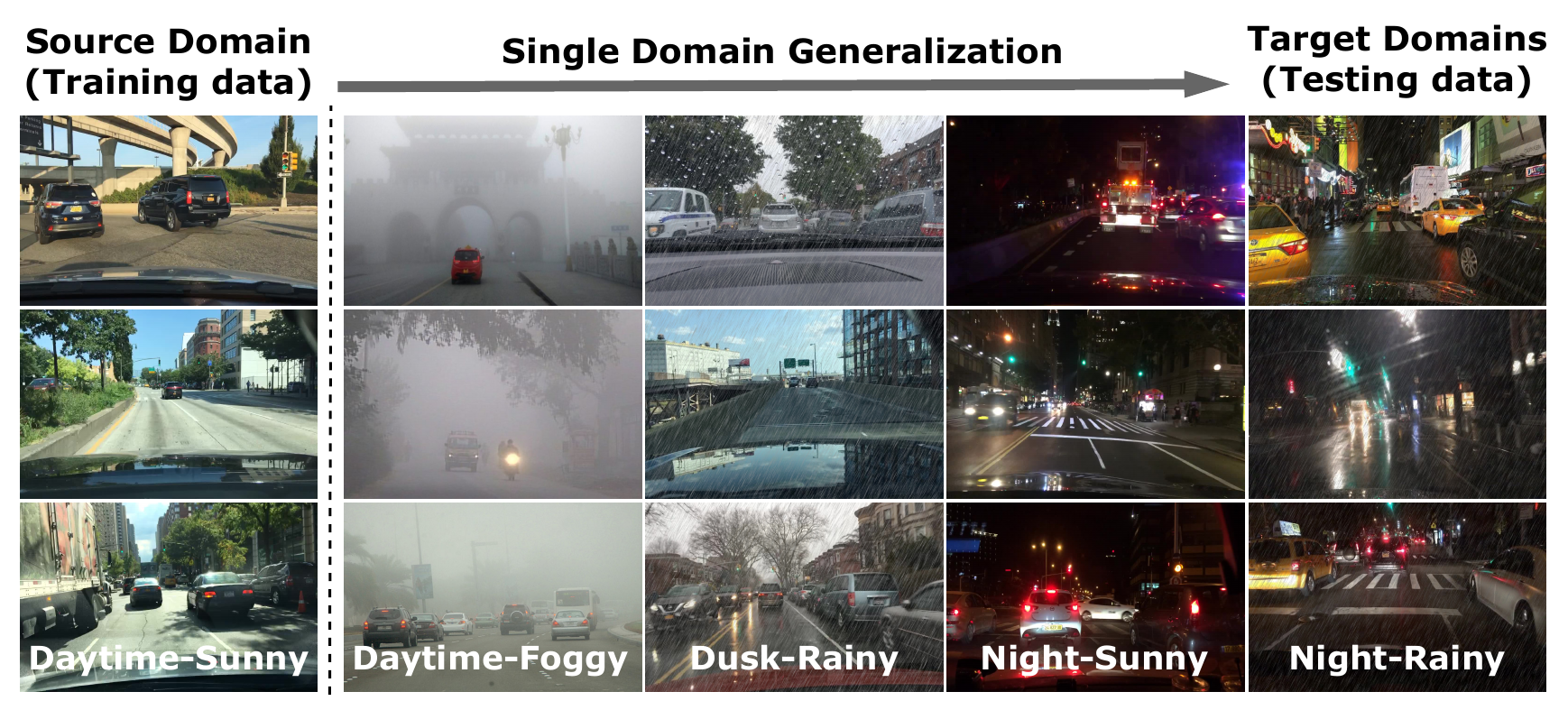}
    \caption{The setting of S-DGOD, which aims to learn from a single source domain and generalize to multiple unseen target domains.
    \wu{It requires extracting the causal features in the source domain for achieving OoD generalization.}}
    \label{fig:sdgod_intro}
\end{figure}

\begin{figure*}[t]
    \centering
    \includegraphics[width=\linewidth]{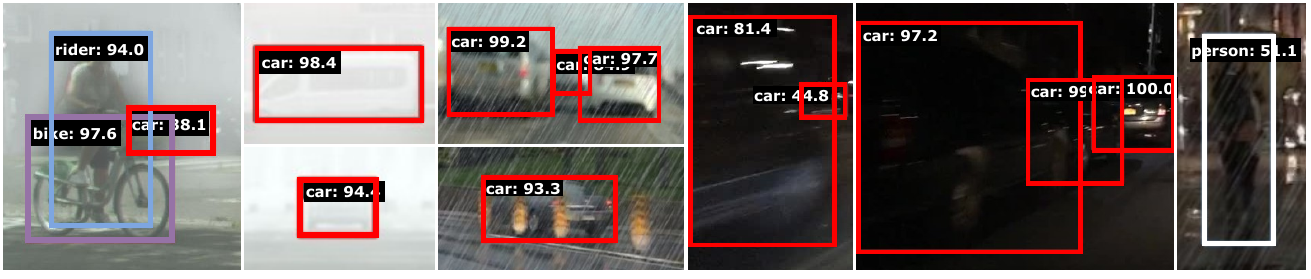}
    \caption{Predictions (category: confidence) of~\methodname~on Single Domain Generalization Object Detection tasks.~\methodname~can detect objects in extremely-challenging environments. Box color indicates the category. Better zoom-in to view.}
    \label{fig:inference_results}
\end{figure*}

Object detection is a fundamental task in computer vision~\cite{ren2015faster, tan2020efficientdet, ge2021yolox, zhang2021varifocalnet}. 
However, improving the generalization ability of object detection remains a challenging problem, especially for Out-of-Distribution scenarios, where data are sampled from novel unseen distributions.
Recently, this has been grounded to a realistic yet challenging task, i.e., Single Domain Generalization Object Detection (S-DGOD)~\cite{wu2022single}, which raised worldwide researchers' attention to the generalization ability of object detection algorithms. The objective of S-DGOD is to improve object detectors' Out-of-Domain (OoD) generalization ability, given a single source domain for training (see~\Cref{fig:sdgod_intro} for illustration).
It requires methods to extract the causal features in the source domain and learn from them for generalization.
Compared with the traditional Domain Generalization (DG), S-DGOD provides only one source domain data, making it easy to over-fit as we are unable to learn the features shared by multiple source domains, which usually contain causal information~\cite{arjovsky2019invariant}.
Most existing works on S-DGOD~\cite{pan2018two, pan2019switchable, huang2019iterative, choi2021robustnet} apply feature normalization to solve the single-domain generalization problem.
Other methods, such as feature disentanglement~\cite{wu2022single} and invariant-based algorithms~\cite{rao2023srcd}, have been proposed.
However, none of these existing works discover the high capacity of architectural design in learning complex data distribution. 
Additionally, there are researches~\cite{ganin2016domain, chen2018domain, saito2019strong, hsu2020progressive, chen2020harmonizing} on \wu{improving object detectors' generalization ability via the Domain Adaption (DA) setting}, which learns from a source domain and generalizes to a specific target domain.
Compared with DA, S-DGOD targets multiple unseen domains, while DA algorithms solely focus on one target domain and \wu{have privileged access} to unlabeled target domain data during training, which makes S-DGOD much more challenging than DA.

In this paper, we propose to leverage the high capacity in fitting complex data of Differentiable NAS to solve the challenging S-DGOD.
Here comes the question: Differentiable NAS methods~\cite{liu2018darts, yang2020ista, zhong2020representation} are known to easily over-fit the training data, how to make them generalizable? DNNs' parameters optimized by gradient descent tend to prioritize learning and making predictions based on easy-to-learn features~\cite{arjovsky2019invariant, jacot2018neural, pezeshki2021gradient}. This phenomenon has further implications on Differentiable NAS, which applies gradient descent to learn the optimal architectural parameters, leading to a bias towards optimizing architecture with a few easy-to-learn features.
Consequently, Differentiable NAS is significantly affected by the spurious correlations between easy features and labels, and easily over-fits the training data, suffering from sub-optimal OoD performance. To address this issue, we propose Generalizable loss (G-loss), which is OoD-aware, to guide the NAS framework and activate both the network's parameters and architectural parameters to learn from not only the easy features but also the remaining predictive features.
\Cref{fig:inference_results} shows the superior performance of our proposed~\methodname.

\wu{
Our main contributions can be summarized as follows:
\begin{itemize}
    \item To the best of our knowledge, our work is the first attempt to introduce Differentiable NAS for S-DGOD, leveraging the high capacity of NAS methods in fitting complex data features.
    \item We propose an OoD-aware objective, namely G-loss, to avoid the NAS process from the over-fitting issue, thus, improving OoD generalization performance.
    \item Extensive experiments demonstrate our proposed~\methodname~empirically outperforms previous SOTA baselines on the challenging S-DGOD benchmarks.
\end{itemize}
}

\section{Related Works}

\begin{figure*}[t]
    \centering
    \includegraphics[width=\linewidth]{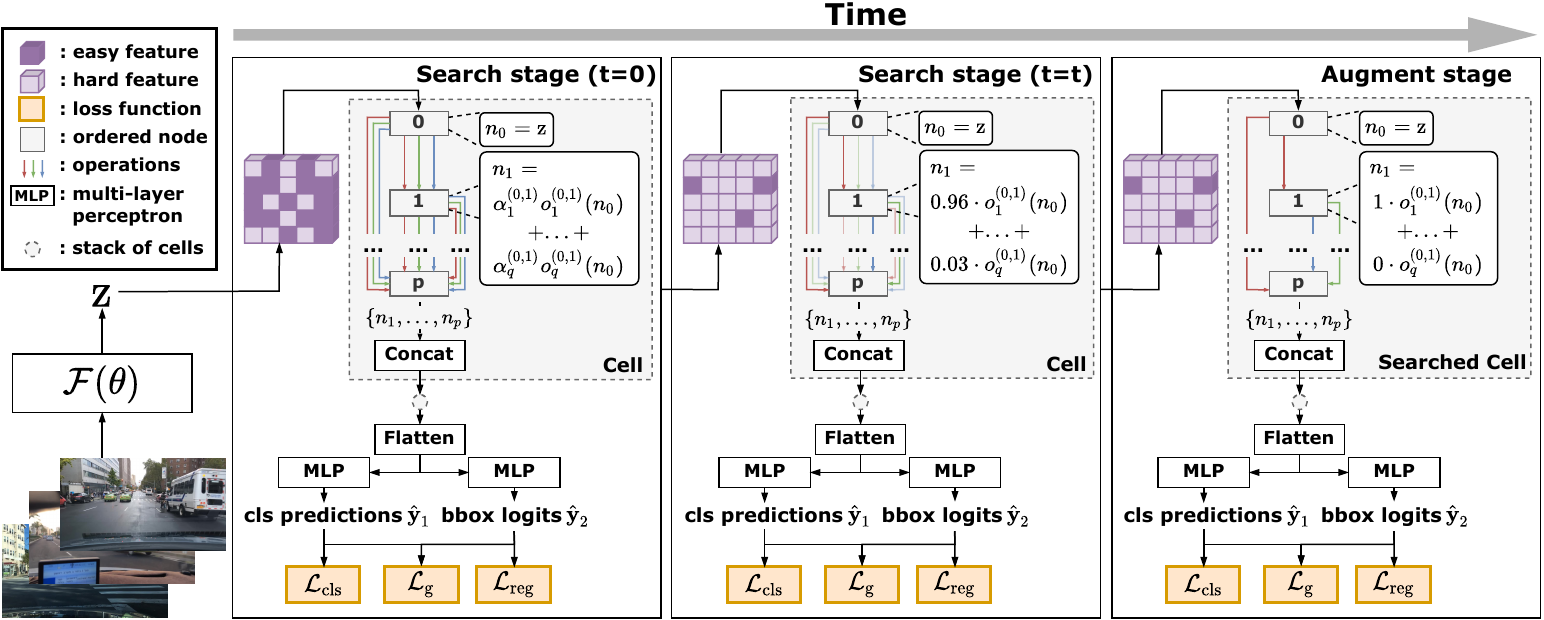}
    \caption{An overview of the proposed~\methodname. 
    At the beginning of the search stage (t=0), the searchable prediction head super-net is randomly initialized, and the feature $\zB$ extracted by the detector network $\Fcal(\theta)$ contains both easy and hard features.
    At the end of the search stage (t=t), the searchable super-net is converged with chosen operation between each node in cells, and the detector network $\Fcal(\theta)$ is forced by $\Lcal_\textnormal{g}$ to learning from hard features, eliminating the influence of the spurious correlation between easy features and ground truth labels.
    At the augment stage, we reconstruct the prediction head with the searched architectural parameters $\alpha^*$ and retrain the whole network.}
    \label{fig:algorithm_framework}
\end{figure*}

\subsection{Single Domain Generalization Object Detection}
The majority of works on S-DGOD can be categorized into two main approaches: feature normalization and invariant-based algorithms.
IBN-Net~\cite{pan2018two} integrates Instance Normalization (IN) and Batch Normalization (BN) into popular deep neural networks to enhance generalization capacity.
Switchable Whitening (SW)~\cite{pan2019switchable} proposes an approach, which selects appropriate whitening methods to adapt to different tasks.
Iterative Normalization (IterNorm)~\cite{huang2019iterative} employs Newton's iterations to efficiently perform feature normalization.
RobustNet (ISW)~\cite{choi2021robustnet} proposes an instance selective whitening loss to disentangle the domain-specific style and domain-invariant feature representations.
Cyclic-Disentangled Self-Distillation (CDSD)~\cite{wu2022single} aims to disentangle domain-invariant representations (DIR) from domain-specific representations and make predictions based on DIR.
Style-Hallucinated Dual Consistency Learning (SHADE)~\cite{zhao2022style} proposes two constraints to encourage models to learn from style-diversified samples while keeping them from over-fitting.
CLIPGap~\cite{vidit2023clip} leverages the pre-trained knowledge of Vision-Language Models (VLM) to enhance the generalization ability.
SRCD~\cite{rao2023srcd} proposes two novel modules to eliminate the effects of spurious correlation and force model to learn from the semantic relationships.
Despite the significant improvements achieved by these works, none of them have explored the potential of NAS algorithms to enhance the capacity of DNNs in fitting complex data distributions.

\subsection{Neural Architecture Search for Object Detection}

Compared with NAS works for the standard image classification tasks, the works of NAS for Object Detection are relatively rare due to their intricacy.
Existing works on NAS for Object Detection can be generally divided into three genres according to the searched component in networks, including backbone search~\cite{cai2018proxylessnas, chen2019detnas, guo2020single, jiang2020sp}, Feature Pyramid Network (FPN) search~\cite{ghiasi2019fpn, liang2021opanas}, and joint detection head and FPN search~\cite{xu2019auto, wang2020fcos}. 
In the setting of Single-DGOD, these works may end up with sub-optimal OoD generalization performance since they aim to find architectures by minimizing the in-distribution \wu{loss}.
On the contrary, our proposed method is guided by an additional OoD-aware objective, enabling us to identify the architecture with optimal OoD performance.

\section{Methodology}

\subsection{Preliminary on Differentiable NAS}
\label{subsec:differentiable_nas}

In this paper, we conduct the Differentiable NAS on the prediction head as shown in~\Cref{fig:algorithm_framework}.
Conventional Differentiable NAS \cite{liu2018darts} aims at utilizing a gradient-based differentiable optimization to search the optimal sub-architecture of the super-net. 
The super-net $\Scal(\omega, \alpha)$ is stacked by several cells which are the computation units to be searched during the training process and are formed as a directed acyclic graph (DAG). 
\wu{There are two types of cells, including normal cells and reduction cells, where the difference is whether the feature maps are down-sampled or not.}
A cell is comprised of $p$ ordered nodes $\Ncal = \{n_1, \dots, n_p\}$ with $q$ candidate operations $\Ocal = \{o_1, \dots, o_q \}$ between each node. 
Binary variables $\delta_{k}^{(i, j)}\in \{0,1\}$ represents whether candidate operation $o_k^{(i, j)}$ between node $n_i$ and $n_j$ is chosen or not. 
Thus, we have the following formulations for each node:
\begin{equation}
    \label{eq:node}
    n_j = \sum_{i=0}^{j-1} \sum_{k=1}^q \delta_k^{(i, j)} o_k^{(i, j)}(n_i) = \bm{\delta}_j^T \textbf{o}_j , 
\end{equation}
where $\bm{\delta}_j^T$ and $\textbf{o}_j$ are vectors formed by $\delta_k^{(i, j)}$ and $o_k(n_i)$ respectively. 
\wu{$\Fcal$ denotes the object detector with network parameters $\theta$, and $\zB$ denotes the feature representations extracted by $\Fcal(\theta)$.}
We use $\zB$ to initialize $n_0 = \zB$ for the first cell and use the output of the previous cell for the rest of the cells.
Practically, DARTS-based \cite{liu2018darts} methods convert $\delta_{k}^{(i, j)}$ into continuous relaxation with a soft-max function to make it differentiable:
\begin{align}
    \label{eq:continuous_relax}
    \alpha_k^{(i, j)} &= \exp \left( \delta_k^{(i, j)} \right) / \sum_k \exp \left( \delta_k^{(i, j)} \right) , \\
    n_j &= \sum_{i=0}^{j-1} \sum_{k=1}^q \alpha_k^{(i, j)} o_k^{(i, j)}(n_i) = \bm{\alpha}_j^T \textbf{o}_j ,
\end{align}
where $\alpha_k^{(i, j)}$ are differentiable and the NAS problem is formulated as the following bi-level optimization problem:
\begin{align}
    \label{eq:bilevel_optimize}
    &\omega^* = \mathop{\arg\min}\limits_\omega \Lcal_\textnormal{train} \left( \Scal \left( \zB; \omega, \alpha \right) \right) , \\
    &\alpha^* = \mathop{\arg\min}\limits_\alpha \Lcal_\textnormal{val} \left( \Scal \left( \zB; \omega^*, \alpha \right) \right) , \\
    &\textnormal{s.t.}~\Vert \alpha_j \Vert_0 = 1, 1 \leq j \leq p ,
\end{align}
\wu{where $\omega$ is the parameters of the prediction head, $\Lcal_\textnormal{train}$ and $\Lcal_\textnormal{val}$ are the training loss and validation loss, respectively.}
During searching process, $\Lcal_\textnormal{train}$ and $\Lcal_\textnormal{val}$ are optimized alternately~\cite{liu2018darts}.
In this paper, we use $\Lcal_\textnormal{train}$ to optimize $\alpha$ as the \wu{in-domain (i.d.) validation set is not suitable for S-DGOD} as we aim to improve OoD generalization ability instead of selecting models with optimal i.d. performance.
\wu{When we get $\alpha^*$, the index of the maximum value in $\alpha^{*(i,j)} \in \Rbb^q, 1 \leq i < j \leq p$ is the chosen operation, then we reconstruct the prediction head and retrain the whole network.}
For the design of the search space, please refer to Appendix.

\subsection{Generalizable Objective}

\begin{figure}[t]
    \centering
    \includegraphics[width=\linewidth]{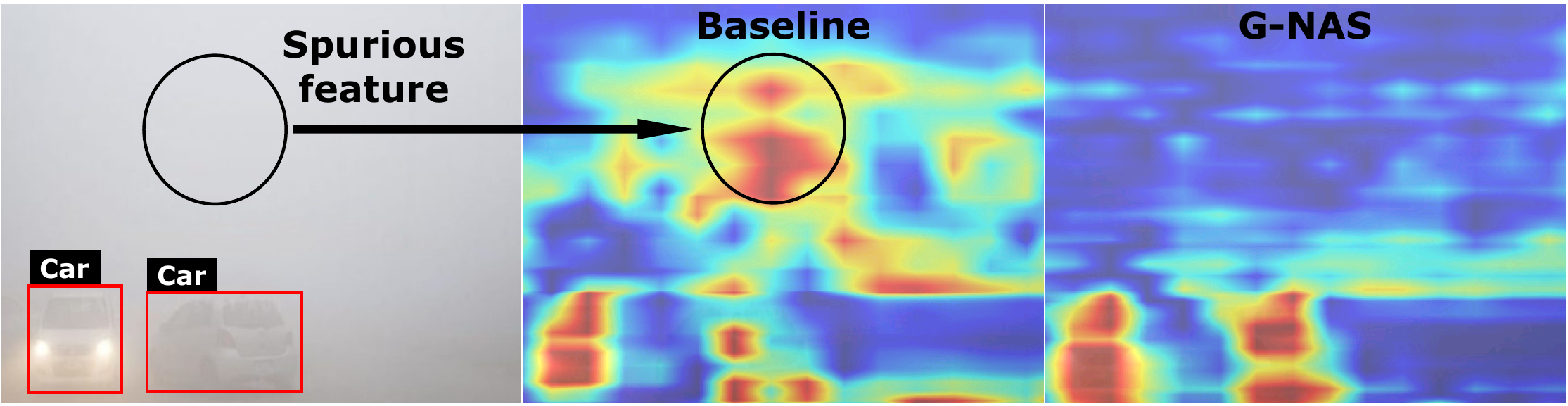}
    \caption{GradCam visualizations on the Daytime-Foggy test set. The results show that the background features significantly mislead the predictions of baseline DNNs, while~\methodname~learns the object-related features to make predictions.}
    \label{fig:spurious_correlation}
\end{figure}

As revealed by~\citet{arjovsky2019invariant},~\citet{jacot2018neural}, and~\citet{pezeshki2021gradient}, DNNs' parameters optimized by gradient descent exhibit an inclination to learn and make predictions based on the easy-to-learn features. 
\wu{These easy-to-learn features are typically non-causal, such as color blocks. For example, as the GradCam maps~\cite{selvaraju2017grad} shown in~\Cref{fig:spurious_correlation},
DNNs could be misled by large salient white blocks and generate false car detection with high confidence. This origins from the spurious correlation between easy-to-learn features (color blocks) and ground truth labels (car annotations) in the daytime training set.}
Consequently, the remaining features that might have causal correlations with the ground truth labels are disregarded. Especially in the setting of S-DGOD, the OoD generalization is hardly achieved as the spurious correlation between easy features and labels learned in the source domain may not exist in the target domains, where the domain gap between source and target domains is huge. This further hinders differentiable NAS and leads to the inclination that only a subset of architectural choices are activated ignoring the remaining architectures, which may possess more significant generalization capabilities in object detection, and resulting in over-fitting.
To address this issue, we propose an OoD-aware objective, Gerneralizable loss (G-loss). 
The goal of G-loss is to discourage using few dominant network's parameters and architectural candidates to make predictions during training, forcing the DNNs and Differentiable NAS to use more abundant information in representation learning.
\wu{
G-loss takes into account the regression branch of the detection network and architectural parameters to regularize the training process.
Our calculation in Theorem~\ref{thm:effectiveness_gloss} leads to the following compact form of G-loss:}
\begin{align}
    \label{eq:g_loss}
    \Lcal_\textnormal{g}(\theta, \omega, \alpha) &= \frac{1}{2} \Vert \hat{\yB}_1 \Vert^2 - \frac{1}{2} \Vert \hat{\yB}_2 \Vert^2 ,
\end{align}
where $\hat{\yB}_1$ and $\hat{\yB}_2$ are the outputs of the classification head and the regression head, respectively.
\wu{We now discuss how G-loss operates to promote balance training.
We assume the width of DNNs goes infinite, in the regime of Neural Tangent Kernel (NTK) theory~\cite{jacot2018neural}, we have the following proposition:}
\begin{restatable}[]{proposition}{lf}
    \label{def:linear_function}
    \textbf{(NTRF approximation of DNNs.)}
    When the width of neural networks goes infinite, the output of over-parameterized neural networks can be approximated as a linear function:
    \begin{align}
        \hat{\yB}_1 = \psi \cdot \Theta \cdot w_1 , \\
        \hat{\yB}_2 = \psi \cdot \Theta \cdot w_2 ,
    \end{align}
    where $\psi \in \Rbb^{n \times m}$ is the Neural Tangent Random Feature (NTRF) matrix~\cite{cao2019generalization} of $n$ training data, 
    $\Theta \in \Rbb^{m}$ denotes the concatenation of all vectorized trainable parameters with size $m$, $w_1 \in \Rbb$ and $w_2 \in \Rbb$ project features into classification output $\hat{\yB}_1 \in \Rbb^{n}$ and regression output $\hat{\yB}_2 \in \Rbb^{n}$, respectively.
\end{restatable}
\wu{Based on this proposition, we have the following theorem:}
\begin{restatable}[]{theorem}{eg}
    \label{thm:effectiveness_gloss}
    Assume the width of the neural network goes infinite.
    We consider G-loss regularized regression loss $\Lcal_\textnormal{reg}$ and classification loss $\Lcal_\textnormal{cls}$:
    \begin{equation}
        \Lcal(\Theta) = \Lcal_\textnormal{cls} + \Lcal_\textnormal{reg} + \Lcal_\textnormal{g} , 
    \end{equation}
    where we apply cross-entropy function for $\Lcal_\textnormal{cls}$ and smooth L1 function for $\Lcal_\textnormal{reg}$.
    The optimization problem:
    \begin{equation}
        \begin{split}
            \min_\Theta \Lcal(\Theta) &= \bm{1} \cdot \log \left[ 1 + \exp \left( - \bm{Y}_1 \hat{\yB}_1 \right) \right] \\
            &+ \frac{1}{2} \left( \hat{\yB}_2 - \yB_2 \right)^T \left( \hat{\yB}_2 - \yB_2 \right) \\
            &+ \frac{1}{2} \Vert \hat{\yB}_1 \Vert^2 - \frac{1}{2} \Vert \hat{\yB}_2 \Vert^2 ,
        \end{split}
    \end{equation}
    where the last two terms are $\Lcal_\textnormal{g}$, $\bm{Y}_1 = \textnormal{diag} (\yB_1) \in \Rbb^{n \times n}$ is the diagonal matrix of ground truth classification labels $\yB_1 \in \Rbb^n$, and $\yB_2 \in \Rbb^n$ is the ground truth regression labels \footnote{To simplify, we consider each image contains one bounding box and one can easily expand it to multiple bounding boxes following the proof of this theorem.}.
    $\bm{1}$ denotes the all-ones vector with size $n$.
    We only consider the interval $\left[-1, 1 \right]$ for smooth L1 function \footnote{We can easily use the normalization to constrain the input of smooth L1 function.}. The above optimization problem can be transferred to the following maximization problem on the dual variable:
    \begin{align}
        &~~~~~~~~~~~~\min_\Theta \Lcal(\Theta) = \max_\Phi \Hcal(\Phi) , \\
        \begin{split}
            \Hcal(\Phi) &= - \bm{1} \cdot \left[ \Phi \log \Phi + (1 - \Phi) \log (1 - \Phi) \right] \\
            &- \Phi^T \bm{Y}_1 \psi \Delta w_1 + \frac{1}{2} \left( \psi \Delta w_2 - \yB_2 \right)^T \left( \psi \Delta w_2 - \yB_2 \right) \\
            &+ \frac{1}{2} \Vert \psi \Delta w_1 \Vert^2 - \frac{1}{2} \Vert \psi \Delta w_2 \Vert^2 ,
        \end{split} \\
        &~~~~~~~~~~\Delta = \psi^{-1} \left( \frac{1}{w_1} \bm{Y}_1^T \Phi + \frac{w_2}{w_1^2} \yB_2 \right) ,
    \end{align}
    where $\Phi \in (0,1)^n$ is a variational parameter defined for each training example.
    Together with~\Cref{def:linear_function}, the gradient descent for $\Phi$ is calculated as follows:
    \begin{equation}
        \label{eq:gradient_descent_phi}
        \frac{\partial \Hcal (\Phi)}{\partial \Phi} = \log \frac{1 - \Phi}{\Phi} - \bm{Y}_1 \bm{Y}_1 ^T \Phi - \frac{w_2}{w_1} \bm{Y}_1 \yB_2 .
    \end{equation}
\end{restatable}
\wu{
Note that $\bm{Y}_1 \bm{Y}_1 ^T$ is a diagonal matrix, and the overall calculation avoids interference between different elements of $\Phi$ in~\Cref{eq:gradient_descent_phi} during gradient descent, making training data independent between each other, thus, avoiding the appearance of dominant features and encourages activating more features for predicting. 
As a result, the whole network is optimized without bias by easy features, learning from not only easy features but also the remaining features with the additional regularization term defined in~\Cref{eq:g_loss}. 
Proof of~\Cref{thm:effectiveness_gloss} can be found in Appendix. 
}





\begin{figure}[t]
\begin{minipage}{\linewidth}
\begin{algorithm}[H]
    \caption{\papertitle}
    \label{alg:methodology}
    \begin{algorithmic}[1]
        \Require Training set $\Dcal_\textnormal{train}$, generalizable loss weight $\lambda_\textnormal{g}$, learning rate $\beta$.
        \Ensure A neural architecture with optimized parameters $\theta^*, \omega^*$ and $\alpha^*$.
        \State \textcolor{gray}{\texttt{\# Search stage}}
        \State Initialize the detector network $\Fcal(\theta)$ ;
        \State Initialize the searchable prediction head $\Scal(\omega, \alpha)$ ;
        \For{each $(\xB, \yB) \in \Dcal_\textnormal{train}$}
            \State Calculate $\Lcal_\textnormal{g}(\theta, \omega, \alpha)$ according to~\Cref{eq:g_loss} ;
            \State Calculate $\Lcal_\textnormal{train}$ according to~\Cref{eq:training_loss};
            \State Update $\theta, \omega, \alpha$ through $\textnormal{SGD} (\Lcal_\textnormal{train}, \beta)$ algorithm ;
        \EndFor
        \State Save the searched architecture $\alpha^*$ ; 
        \State \textcolor{gray}{\texttt{\# Augment stage}}
        \State Initialize detector network $\Fcal(\theta)$ ;
        \State Reconstruct the searched prediction head $\Scal(\omega, \alpha^*)$ ;
        \For{each $(\xB, \yB) \in \Dcal_\textnormal{train}$}
            \State Calculate $\Lcal_\textnormal{g}(\theta, \omega, \alpha^*)$ according to~\Cref{eq:g_loss} ;
            \State Calculate $\Lcal_\textnormal{train}$ according to~\Cref{eq:training_loss};
            \State Update $\theta, \omega$ through $\textnormal{SGD} (\Lcal_\textnormal{train}, \beta)$ algorithm ;
        \EndFor
        \State Save the optimized parameters $\theta^*, \omega^*$ ; 
    \end{algorithmic}
\end{algorithm}
\end{minipage}
\end{figure}

\subsection{Algorithm Framework}
\label{subsec:algorithm_framework}

Our proposed G-NAS is outlined in~\Cref{alg:methodology} and the structure is depicted in~\Cref{fig:algorithm_framework}.
The whole algorithm is built upon Faster R-CNN~\cite{ren2015faster} and contains two stages: the search stage and the augment stage.
Firstly, in the search stage, a super-net prediction head $\Scal(\omega, \alpha)$ is constructed according to~\Cref{eq:continuous_relax}. 
We repeat to update the trainable parameters using the Stochastic Gradient Descent algorithm.
In the augment stage, we apply the searched architectural parameters $\alpha^*$ to reconstruct the prediction head $\Scal(\omega, \alpha^*)$, where we only construct the chosen operation.
\wu{The visualization of the searched architectures can be found in Appendix.}
For both stages, the training loss $\Lcal_\textnormal{train}$ is calculated as followed:
\begin{equation}
    \label{eq:training_loss}
    \Lcal_\textnormal{train} =  \Lcal_\textnormal{det} + \Lcal_\textnormal{cls} + \Lcal_\textnormal{reg} + \lambda_g \cdot \Lcal_g ,
\end{equation}
where $\Lcal_\textnormal{det}$ is the detector loss~\cite{ren2015faster} for the region proposal network (RPN), $\Lcal_\textnormal{cls}$ is the cross-entropy loss used for classification, $\Lcal_\textnormal{reg}$ is the smooth L1 loss used for bounding box regression, and $\lambda_\textnormal{g}$ is a hyper-parameter to determine the weight of $\Lcal_\textnormal{g}$.


\begin{table*}[t]
    \centering
    \begin{threeparttable}
        \begin{tabularx}{\linewidth}{X|c|cccc|c}
            \toprule
            \textbf{Method} & \textbf{Daytime-Sunny} & \textbf{Daytime-Foggy} & \textbf{Dusk-Rainy} & \textbf{Night-Sunny} & \textbf{Night-Rainy} & \textbf{Average} \\
            \midrule
            Faster R-CNN~(\citeyear{ren2015faster}) & 51.1 & 31.9 & 26.6 & 33.5 & 14.5 & 26.6 \\
            IBN-Net~(\citeyear{pan2018two}) & 49.7 & 29.6 & 26.1 & 32.1 & 14.3 & 25.5 \\
            SW~(\citeyear{pan2019switchable}) & 50.6 & 30.8 & 26.3 & 33.4 & 13.7 & 26.1 \\
            IterNorm~(\citeyear{huang2019iterative}) & 43.9 & 28.4 & 22.8 & 29.6 & 12.6 & 23.4 \\
            ISW~(\citeyear{choi2021robustnet}) & 51.3 & 31.8 & 25.9 & 33.2 & 14.1 & 26.3 \\
            CDSD~(\citeyear{wu2022single}) & \underline{56.1} & 33.5 & 28.2 & 36.6 & 16.6 & 28.7 \\
            SHADE~(\citeyear{zhao2022style}) & - & 33.4 & \underline{29.5} & 33.9 & 16.8 & 28.4 \\
            CLIPGap~(\citeyear{vidit2023clip}) & 48.1 & 32.0 & 26.0 & 34.4 & 12.4 & 26.2 \\
            SRCD~(\citeyear{rao2023srcd}) & - & \underline{35.9} & 28.8 & \underline{36.7} & \underline{17.0} & \underline{29.6} \\
            \midrule
            \textbf{\methodname}~(Ours) & \textbf{58.4} & \textbf{36.4} & \textbf{35.1} & \textbf{45.0} & \textbf{17.4} & \textbf{33.5} \\
            \bottomrule
        \end{tabularx}
    \end{threeparttable}
    \caption{
    Single domain generalization object detection results.
    All algorithms are trained on the Daytime-Sunny domain and tested on the other four domains.
    Average results are calculated using four out-of-domain results to compare the Out-of-Domain generalization ability.
    All baseline results are mainly taken from SHADE~\cite{zhao2022style}, CLIPGap~\cite{vidit2023clip}, and SRCD~\cite{rao2023srcd}.
    ``-'' denotes the results without reporting in the original paper.
    The numbers in bold and underlined denote the highest and the second performance, respectively.
    The results demonstrate that our approach is robust against domain shifts and achieves the SOTA S-DGOD performance.}
    \label{tab:single_domain_generalization}
\end{table*}



\begin{table*}[t]
    \centering
    \begin{adjustbox}{width=\linewidth}
	\begin{threeparttable}
        \begin{tabular}{l|ccccccc|ccccccc}
            \toprule
            & \multicolumn{7}{c|}{\textbf{Daytime-Foggy}} & \multicolumn{7}{c}{\textbf{Dusk-Rainy}} \\
            \textbf{Method} & \textbf{Bus} & \textbf{Bike} & \textbf{Car} & \textbf{Motor} & \textbf{Person} & \textbf{Rider} & \textbf{Truck} & \textbf{Bus} & \textbf{Bike} & \textbf{Car} & \textbf{Motor} & \textbf{Person} & \textbf{Rider} & \textbf{Truck} \\
            \midrule
            FR & 30.7 & 26.7 & 49.7 & 26.2 & 30.9 & 35.5 & 23.2 & 36.8 & 15.8 & 50.1 & 12.8 & 18.9 & 12.4 & 39.5 \\
            IBN-Net & 29.9 & 26.1 & 44.5 & 24.4 & 26.2 & 33.5 & 22.4 & 37.0 & 14.8 & 50.3 & 11.4 & 17.3 & 13.3 & 38.4 \\
            SW & 30.6 & 26.2 & 44.6 & 25.1 & 30.7 & 34.6 & 23.6 & 35.2 & 16.7 & 50.1 & 10.4 & \underline{20.1} & 13.0 & 38.8 \\
            IterNorm & 29.7 & 21.8 & 42.4 & 24.4 & 26.0 & 33.3 & 21.6 & 32.9 & 14.1 & 38.9 & 11.0 & 15.5 & 11.6 & 35.7 \\
            ISW & 29.5 & 26.4 & 49.2 & 27.9 & 30.7 & 34.8 & 24.0 & 34.7 & 16.0 & 50.0 & 11.1 & 17.8 & 12.6 & 38.8 \\
            CDSD & \underline{32.9} & 28.0 & 48.8 & 29.8 & 32.5 & 38.2 & 24.1 & 37.1 & 19.6 & \underline{50.9} & \underline{13.4} & 19.7 & 16.3 & \underline{40.7} \\
            SRCD & \textbf{36.4} & \underline{30.1} & \underline{52.4} & \underline{31.3} & \underline{33.4} & \textbf{40.1} & \textbf{27.7} & \underline{39.5} & \underline{21.4} & 50.6 & 11.9 & \underline{20.1} & \underline{17.6} & 40.5 \\
            \midrule
            \textbf{\methodname} & 32.4 & \textbf{31.2} & \textbf{57.7} & \textbf{31.9} & \textbf{38.6} & \underline{38.5} & \underline{24.5} & \textbf{44.6} & \textbf{22.3} & \textbf{66.4} & \textbf{14.7} & \textbf{32.1} & \textbf{19.6} & \textbf{45.8} \\
            \bottomrule
        \end{tabular}
        \end{threeparttable}
    \end{adjustbox}
    \caption{Per-class results on Daytime-Foggy and Dusk-Rainy.
    FR denotes Faster R-CNN.
    }
    \label{tab:df_dr}
\end{table*}


\begin{table*}[t]
    \centering
	\begin{adjustbox}{width=\linewidth}
        \begin{tabular}{l|ccccccc|ccccccc}
            \toprule
            & \multicolumn{7}{c|}{\textbf{Night-Sunny}} & \multicolumn{7}{c}{\textbf{Night-Rainy}} \\
            \textbf{Method} & \textbf{Bus} & \textbf{Bike} & \textbf{Car} & \textbf{Motor} & \textbf{Person} & \textbf{Rider} & \textbf{Truck} & \textbf{Bus} & \textbf{Bike} & \textbf{Car} & \textbf{Motor} & \textbf{Person} & \textbf{Rider} & \textbf{Truck} \\
            \midrule
            FR & 37.7 & 30.6 & 49.5 & 15.4 & 31.5 & 28.6 & 40.8 & 22.6 & 11.5 & 27.7 & 0.4 & 10.0 & 10.5 & 19.0 \\
            IBN-Net & 37.8 & 27.3 & 49.6 & 15.1 & 29.2 & 27.1 & 38.9 & 24.6 & 10.0 & 28.4 & \underline{0.9} & 8.3 & 9.8 & 18.1\\
            SW & 38.7 & 29.2 & 49.8 & 16.6 & 31.5 & 28.0 & 40.2 & 22.3 & 7.8 & 27.6 & 0.2 & 10.3 & 10.0 & 17.7 \\
            IterNorm & 38.5 & 23.5 & 38.9 & 15.8 & 26.6 & 25.9 & 38.1 & 21.4 & 6.7 & 22.0 & \underline{0.9} & 9.1 & 10.6 & 17.6 \\
            ISW & 38.5 & 28.5 & 49.6 & 15.4 & 31.9 & 27.5 & 41.3 & 22.5 & 11.4 & 26.9 & 0.4 & 9.9 & 9.8 & 17.5 \\
            CDSD & 40.6 & \underline{35.1} & 50.7 & 19.7 & 34.7 & \underline{32.1} & \underline{43.4} & 24.4 & \underline{11.6} & 29.5 & \textbf{9.8} & \underline{10.5} & \underline{11.4} & 19.2 \\
            SRCD & \underline{43.1} & 32.5 & \underline{52.3} & \underline{20.1} & \underline{34.8} & 31.5 & 42.9 & \underline{26.5} & \textbf{12.9} & \underline{32.4} & 0.8 & 10.2 & \textbf{12.5} & \textbf{24.0} \\
            \midrule
            \textbf{\methodname} & \textbf{46.9} & \textbf{40.5} & \textbf{67.5} & \textbf{26.5} & \textbf{50.7} & \textbf{35.4} & \textbf{47.8} & \textbf{28.6} & 9.8 & \textbf{38.4} & 0.1 & \textbf{13.8} & 9.8 & \underline{21.4} \\
            \bottomrule
        \end{tabular}
    \end{adjustbox}
    \caption{Per-class results on Night-Sunny and Night-Rainy.
    }
    \label{tab:nc_ns}
\end{table*}

\section{Experiments}
\label{sec:experiments}


\subsection{Experimental Setup}
\label{subsec:experimental_setup}

\paragraph{Datasets.} 
To evaluate different methods' single-domain generalization ability, we follow the setting proposed by~\citet{wu2022single}.
The dataset contains five urban-scene domains with distinct weather conditions, including Daytime-Sunny, Daytime-Foggy, Dusk-Rainy, Night-Sunny, and Night-Rainy.
The Daytime-Sunny is the source training domain and the other four domains are \wu{only} used for testing.
More details about the construction of these domains can be found in Appendix.


\paragraph{Baselines.}
We choose eight classic and SOTA algorithms from the Single-DGOD benchmarks~\cite{wu2022single} for comparison, including IBN-Net~\cite{pan2018two}, Switchable Whitening (SW)~\cite{pan2019switchable}, Iterative Normalization (IterNorm)~\cite{huang2019iterative}, RobustNet (ISW)~\cite{choi2021robustnet}, Cyclic-Disentangled Self-Distillation (CDSD)~\cite{wu2022single}, Style-Hallucinated Dual Consistency Learning (SHADE)~\cite{zhao2022style}, CLIPGap~\cite{vidit2023clip}, and SRCD~\cite{rao2023srcd}.
For CLIPGap, we use the version initialized with the ImageNet pre-trained weights for fair comparisons as all other baseline methods solely apply ImageNet pre-training.

\paragraph{Evaluation Metric.}
For all quantitative experiments, we follow~\citet{wu2022single} to evaluate methods' performance using Mean Average Precision (mAP) and report the AP of each class, which is known as PASCAL VOC evaluation metric~\cite{everingham2010pascal}.

\paragraph{Implemetation Details.}
We apply the Faster R-CNN detector~\cite{ren2015faster} with ResNet-101 backbone~\cite{he2016deep} as the base model for all algorithms to perform object detection.
All algorithms are solely initialized by the ImageNet pre-trained weights.
We replace the prediction head with our proposed NAS framework, which contains a stem convolution layer, a searchable normal cell, and a searchable reduction cell.
The whole training process consists of two sequentially executing stages, the search stage, and the augment stage.
We first construct a searchable super-net in the search stage to perform the architecture search within the super-net and save the architecture searched in the last epoch.
In the second stage, we reconstruct the prediction head with the architectural parameters obtained in the first stage and perform end-to-end training.
\wu{We train all models until full convergence for 12 epochs.}
We set the $\lambda_\textnormal{g}$ to 1.0.
All parameters in our NAS framework are randomly initialized. 
We apply an SGD optimizer with the learning rate set to 0.02 and we set the batch size to 4 per GPU. 
All experiments are conducted on a computer with 8 GPUs.

\subsection{Comparison with the State of the Art}
\paragraph{Overall Single-DGOD Results.}
\Cref{tab:single_domain_generalization} shows the mAP results on all domains, including Daytime-Sunny, Daytime-Foggy, Dusk-Rainy, Night-Sunny, and Night-Rainy, in which Daytime-Sunny is for training. The average mAPs on test domains are also reported to measure generalization abilities.
As shown in \Cref{tab:single_domain_generalization}, our proposed~\methodname~significantly improves the average generalization performance to 33.5\% compared with the SOTA algorithm, \ie, SRCD, which achieves 29.6\%.
Notably,~\methodname~simultaneously achieves SOTA performance on all target domains, where the domain gaps between the source domain (Daytime-Sunny) and each of these target domains are large.
This suggests~\methodname~is more generalizable and robust against domain shifts compared with baselines. 

\paragraph{Daytime-Sunny to Daytime-Foggy.}
In the Daytime-Foggy scenario, objects are covered by fog, posing extremely challenging test environments.
\Cref{tab:df_dr} lists per-class results on Daytime-Foggy.
Notably,~\methodname~brings the AP of person class up to 38.6\%, surpassing baselines by 5.2\%.
Compared to other objects, the person class object is generally smaller and much more difficult to detect under adverse weather conditions, 
\wu{while in autonomous driving, it is crucial to accurately detect pedestrians on roads.} 
As shown in~\Cref{fig:spurious_correlation}, G-NAS solves this problem by avoiding fitting spurious correlations that exist in the training data. This result demonstrates that~\methodname~is effective in detecting difficult objects, serving for life-critical applications, such as autonomous driving.

\paragraph{Daytime-Sunny to Dusk-Rainy.} 
Dusk-Rainy significantly differs from the source domain, not only in the change of the weather but also in the time.
In rainy scenes, vehicles' light will be reflected on objects' surface by the water on the ground, causing changes in objects' appearance and making them harder to identify.
As shown in~\Cref{tab:df_dr}, our method improves APs of most vehicle classes and achieves SOTA performance.

\paragraph{Daytime-Sunny to Night-Sunny.} 
The night scenarios have been challenging in various research as the lighting condition is too bad to clearly identify completed objects.
\wu{Specifically, if DNN learns to predict based on spurious correlations, such as the color of cars, its performance might confront severe degeneration as the color is not salient in the night scenario compared with daytime.}
\Cref{tab:single_domain_generalization} and \Cref{tab:nc_ns} shows that our proposed~\methodname~significantly surpasses SOTA baselines by 8.3\% mAP.
\wu{These results demonstrate the feature representations learned by~\methodname~are more generalizable, overcoming the influence of spurious correlations.}


\begin{table}[t]
    \centering
    \begin{adjustbox}{width=\linewidth}
        \begin{tabular}{l|cc|cccc|c}
            \toprule
            \textbf{Method} & \textbf{NAS} & \textbf{G-loss} & \textbf{D-F} & \textbf{D-R} & \textbf{N-S} & \textbf{N-R} & \textbf{Avg.} \\
            \midrule
            \methodname & \ding{56} & \ding{56} & 32.3 & 27.0 & 33.9 & 14.8 & 27.0 \\
            \methodname & \ding{56} & \ding{52} & 34.6 & 31.9 & 40.7 & 17.0 & 31.1 \\
            \methodname & \ding{52} & \ding{56} & 33.6 & 28.0 & 35.2 & 16.1 & 28.2 \\
            \methodname & \ding{52} & \ding{52} & \textbf{36.4} & \textbf{35.1} & \textbf{45.0} & \textbf{17.4} & \textbf{33.5} \\
            \bottomrule
        \end{tabular}
    \end{adjustbox}
    \caption{
    Ablation study on~\methodname.
    D, F, R, N, and S represent Daytime, Foggy, Rainy, Night, and Sunny, respectively.
    Avg. denotes the average performance on the four domains.
    }
    \label{tab:ablation_study}
\end{table}

\paragraph{Daytime-Sunny to Night-Rainy.} 
Night-Rainy is a challenging scenario as the low-light condition is synergized with the rainy weather.
\Cref{tab:nc_ns} shows our method achieves the best performance in pedestrian detection and our method achieves 17.4\% mAP in \Cref{tab:single_domain_generalization}, outperforming SOTA baselines by 0.4\%.
\wu{These results demonstrate that~\methodname~is robust even in extremely bad conditions.}

\subsection{Ablation Study}

\paragraph{NAS.}
For the experiments without NAS, We randomly initialize the $\alpha$ and fix the architecture during the whole training process.
\Cref{tab:ablation_study} shows that our NAS strategy brings the average performance up to 33.5\% mAP with G-loss and 28.2\% mAP without G-loss, surpassing~\methodname~without NAS by 2.4\% mAP and 1.2\%, respectively.
These results show that our NAS strategy plays a crucial role in improving the generalization performance, demonstrating the architectural design has a significant influence on networks' generalization ability.

\begin{figure}[t]
    \centering
    \includegraphics[width=0.9\linewidth]{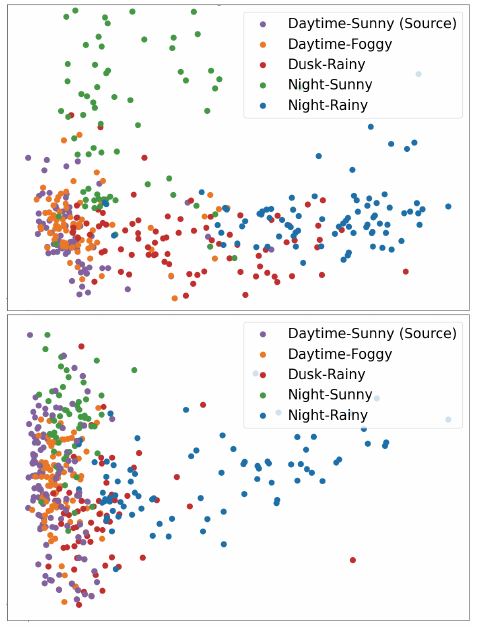}
    \caption{PCA projections of the representations on different domains.
    The feature representations learned with G-loss (bottom) have more similar patterns across different domains than without G-loss (top).
    }
    \label{fig:pca}
\end{figure}

\paragraph{G-loss.}
For the ablation study on G-loss, we simply set the $\lambda_\textnormal{g}$ to 0 to eliminate the influence of G-loss.
As shown in~\Cref{tab:ablation_study}, G-loss brings the average performance up to 33.5\% mAP with NAS and 31.1\% without NAS, surpassing~\methodname~without G-loss by 2.5\% mAP and 4.1\%, respectively.
These results reveal that G-loss is efficient for guiding the NAS framework to find the architecture with optimal generalization performance \wu{by activating the whole network to learn from causal features.
We also conduct an ablation study on the hyper-parameter $\lambda_\textnormal{g}$ in Appendix, and the results show that~\methodname~achieves the best OoD performance when $\lambda_\textnormal{g}$ is set to 1, which aligns well with previous theoretical analysis in~\Cref{thm:effectiveness_gloss}.}

\subsection{Visualization}



\Cref{fig:pca} shows that our NAS framework trained with G-loss extracts similar feature representations on source and target domains, \wu{which indicates the learned features are generalizable and causal as they are stable against domain shifts.}
On the contrary, the NAS framework trained without G-loss over-fits to the Daytime-Sunny domain, \wu{generalizing to the Daytime-Foggy domain but inconsistently performing in the remaining target domains.}


\section{Conclusion}

In this work, we primarily focus on the challenging S-DGOD scenario, which holds substantial real-world significance. S-DGOD involves training object detectors on a single source domain and enabling them to generalize to numerous unseen target domains. To address this challenge, we harness the potent capacity of NAS techniques to model intricate data distributions. Additionally, We introduce an OoD-aware objective, termed G-loss, to augment the NAS framework in learning crucial information. The experimental results highlight that our proposed approach, \methodname, surpasses state-of-the-art methods across four distinct weather conditions in S-DGOD benchmarks. Furthermore, our ablation study underscores the indispensability of each proposed module for achieving robust generalization performance. To the best of our knowledge, this study marks the pioneering attempt to tackle NAS in S-DGOD, yielding state-of-the-art performance concurrently.

\clearpage

\section{Acknowledgments}
Nanyang Ye was supported in part by National Natural Science Foundation of China under Grant No.62106139, 61960206002, 62272301, 62020106005, 62061146002, 62032020, in part by Shanghai Artificial Intelligence Laboratory and the National Key R\&D Program of China under Grant No.2022ZD0160100.


\nocite{*}
\bibliography{aaai24}

\clearpage
\appendix

\section{Appendix}



\subsection{Proof of~\Cref{thm:effectiveness_gloss}}

\noindent We first restate the proposition:
\lf*
\noindent We then restate the theorem:
\eg*

\noindent The proof of~\Cref{thm:effectiveness_gloss} is shown as follows:
\begin{proof}
    Recall the optimization problem:
    \begin{equation}
        \begin{split}
            \min_\Theta \Lcal(\Theta) &= \bm{1} \cdot \log \left[ 1 + \exp \left( - \bm{Y}_1 \hat{\yB}_1 \right) \right] \\
            &+ \frac{1}{2} \left( \hat{\yB}_2 - \yB_2 \right)^T \left( \hat{\yB}_2 - \yB_2 \right) \\
            &+ \frac{1}{2} \Vert \hat{\yB}_1 \Vert^2 - \frac{1}{2} \Vert \hat{\yB}_2 \Vert^2 ,
        \end{split}
    \end{equation}
    According to~\cite{jaakkola1999probabilistic}, we have the following inequality:
    \begin{equation}
        \begin{split}
            \log \left[ 1 + \exp \left( - \bm{Y}_1 \hat{\yB}_1 \right) \right] \geq &- \left[ \Phi \log \Phi + (1 - \Phi) \log (1 - \Phi) \right] \\
            &- \Phi \odot \bm{Y}_1 \hat{\yB}_1 , \\
        \end{split}
    \end{equation}
    where $\odot$ is the element-wise vector product, and the equality holds when $\Phi$ achieves $\Phi^*=\frac{\partial \Lcal}{\partial \bm{Y}_1 \hat{\yB}_1}$~\cite{pezeshki2021gradient}.
    The minimization problem of $\Lcal$ w.r.t $\Theta$ can be written as:
    \begin{align}
        \label{eq:problem_transfer}
        \min_\Theta \Lcal(\Theta) &= \min_\Theta \max_\Phi \Hcal(\Phi, \Theta) , \\
        \begin{split}
            \Hcal(\Phi, \Theta) &= - \bm{1} \cdot \left[ \Phi \log \Phi + (1 - \Phi) \log (1 - \Phi) \right] \\
            &- \Phi^T \bm{Y}_1 \hat{\yB}_1 + \frac{1}{2} \left( \hat{\yB}_2 - \yB_2 \right)^T \left( \hat{\yB}_2 - \yB_2 \right) \\
            &+ \frac{1}{2} \Vert \hat{\yB}_1 \Vert^2 - \frac{1}{2} \Vert \hat{\yB}_2 \Vert^2 .
        \end{split}
    \end{align}
    Note that $\min_\Theta$ and $\max_\Phi$ can be swapped according to Lemma 3 in~\cite{jaakkola1999probabilistic} and we have:
    \begin{equation}
        \begin{split}
            \min_\Theta \Lcal(\Theta) = \max_\Phi \min_\Theta \Hcal(\Phi, \Theta)
        \end{split}
    \end{equation}
    Together with~\Cref{def:linear_function}, we have the solution $\Theta^*$ for the r.h.s.:
    \begin{align}
        &\frac{\partial \Hcal(\Phi, \Theta)}{\partial \Theta} \big|_{\Theta = \Theta^*} = 0 , \\
        \begin{split}
            \frac{\partial \Hcal(\Phi, \Theta)}{\partial \Theta} &= - w_1 \psi^T \bm{Y}_1^T \Phi + w_2 \psi^T \left( \psi \Theta w_2 - \yB_2 \right) \\
            &+ w_1^2 \psi^T \psi \Theta - w_2^2 \psi^T \psi \Theta \\
            &= - w_1 \psi^T \bm{Y}_1^T \Phi + w_2^2 \psi^T \psi \Theta - w_2 \psi^T \yB_2 \\
            &+ w_1^2 \psi^T \psi \Theta - w_2^2 \psi^T \psi \Theta \\
            &= - w_1 \psi^T \bm{Y}_1^T \Phi - w_2 \psi^T \yB_2 + w_1^2 \psi^T \psi \Theta . \\
        \end{split}
    \end{align}
    Then we have:
    \begin{align}
        \Theta^*(\Phi) &= \psi^{-1} \left( \frac{1}{w_1} \bm{Y}_1^T \Phi + \frac{w_2}{w_1^2} \yB_2 \right) , \\
        \Delta &= \Theta^*(\Phi) , \frac{\partial \Delta}{\partial \Phi} = \frac{\psi^{-1} \bm{Y}_1^T}{w_1} .
        \label{eq:gradient_of_theta_wrt_phi}
    \end{align}
    Therefore,~\Cref{eq:problem_transfer} is transferred to
    \begin{align}
        &\min_\Theta \Lcal(\Theta) = \max_\Phi \Hcal(\Phi) , \\
        \begin{split}
            \Hcal(\Phi) &= - \bm{1} \cdot \left[ \Phi \log \Phi + (1 - \Phi) \log (1 - \Phi) \right] \\
            &- \Phi^T \bm{Y}_1 \psi \Delta w_1 + \frac{1}{2} \left( \psi \Delta w_2 - \yB_2 \right)^T \left( \psi \Delta w_2 - \yB_2 \right) \\
            &+ \frac{1}{2} \Vert \psi \Delta w_1 \Vert^2 - \frac{1}{2} \Vert \psi \Delta w_2 \Vert^2 .
        \end{split}
    \end{align}
    Together with~\Cref{eq:gradient_of_theta_wrt_phi}, we have the following deduction:
    \begin{equation}
        \begin{split}
            \frac{\partial \Hcal(\Phi)}{\partial \Phi} &= \log (1 - \Phi) - \log \Phi \\
            &- 2 \bm{Y}_1 \bm{Y}_1 ^T \Phi - \frac{w_2}{w_1} \bm{Y}_1 \yB_2 + \frac{w_2}{w_1} \bm{Y}_1 \left(\psi \Delta w_2 - \yB_2 \right) \\
            &+ \bm{Y}_1 \psi \Delta w_1 - \frac{w_2}{w_1} \bm{Y}_1 \psi \Delta w_2 \\
            \frac{\partial \Hcal(\Phi)}{\partial \Phi} &= \log \frac{1 - \Phi}{\Phi} - 2 \bm{Y}_1 \bm{Y}_1 ^T \Phi -  \frac{2w_2}{w_1} \bm{Y}_1 \yB_2 \\
            &+ w_1 \bm{Y}_1 \psi \psi^{-1} \left( \frac{1}{w_1} \bm{Y}_1^T \Phi + \frac{w_2}{w_1^2} \yB_2 \right) \\
            \frac{\partial \Hcal(\Phi)}{\partial \Phi} &= \log \frac{1 - \Phi}{\Phi} - \bm{Y}_1 \bm{Y}_1 ^T \Phi - \frac{w_2}{w_1} \bm{Y}_1 \yB_2 .
        \end{split}
    \end{equation}
\end{proof}

\subsection{Datasets}
The Daytime-Sunny, Daytime-Foggy, Dusk-Rainy, Night-Sunny and Night-Rainy are constructed using Berkeley Deep Drive 100K (BDD100K)~\cite{yu2020bdd100k}, Cityscapes~\cite{cordts2016cityscapes}, Foggy Cityscapes~\cite{sakaridis2018semantic}, and Adverse-Weather datasets~\cite{hassaballah2020vehicle}.
Particularly, the Daytime-Sunny domain contains 27,708 images in total (19,395 images for training and 8,313 images for testing) selected from the BDD100K dataset.
Specifically, the Daytime-Foggy domain contains 3,775 images selected from Foggy Cityscapes and Adverse-Weather datasets.
The Dusk-Rainy and Night-Rainy domains contain 3,501 and 2,494 images, respectively, rendered from the BDD100K dataset following~\citet{wu2021vector}.
The Night-Sunny domain contains 26,158 images selected from the BDD100K dataset.
\wu{For consistency, we follow~\citet{wu2022single} and consider seven common categories, including bus, bike, car, motor, person, rider, and truck.}

\subsection{Further Experiments}

We conduct an ablation study on the value of $\lambda_\textnormal{g}$ to understand how it affects the overall generalization performance and find the optimal setting for it.
\Cref{tab:hyper_parameters} lists the results of different values for $\lambda_\textnormal{g}$ on the four target domains.
These results show that~\methodname~achieves the optimal average generalization performance with up to 33.5\% when $\lambda_\textnormal{g}$ is set to $1$.
This is consistent with the value of $\lambda_\textnormal{g}$ set in~\Cref{thm:effectiveness_gloss} and further validate \Cref{thm:effectiveness_gloss}.


\begin{table}[ht]
    \centering
    \begin{adjustbox}{width=\linewidth}
        \begin{tabular}{l|c|cccc|c}
            \toprule
            \textbf{Method} & $\lambda_\textnormal{g}$ & \textbf{D-F} & \textbf{D-R} & \textbf{N-S} & \textbf{N-R} & \textbf{Avg.} \\
            \midrule
            \methodname & 0 & 33.6 & 28.0 & 35.2 & 16.1 & 28.2 \\
            \methodname & 0.01 & 34.0 & 31.2 & 41.4 & \textbf{17.5} & 31.0 \\
            \methodname & 0.1 & 34.1 & 31.5 & 41.6 & 17.0 & 31.1 \\
            \methodname & 1.0 & \textbf{36.4} & \textbf{35.1} & \textbf{45.0} & 17.4 & \textbf{33.5} \\
            \methodname & 2.0 & 34.6 & 31.8 & 42.0 & 17.3 & 31.4 \\
            \methodname & 5.0 & 34.3 & 31.0 & 40.3 & 16.9 & 30.6 \\
            \methodname & 10.0 & 32.3 & 29.5 & 39.3 & 16.7 & 29.5 \\
            \bottomrule
        \end{tabular}
    \end{adjustbox}
    \caption{
    \textbf{Results of different hyper-parameters.}
    D, F, R, N, and S represent Daytime, Foggy, Rainy, Night, and Sunny, respectively.}
    \label{tab:hyper_parameters}
\end{table}

\begin{figure}[t]
    \centering
    \includegraphics[width=\linewidth]{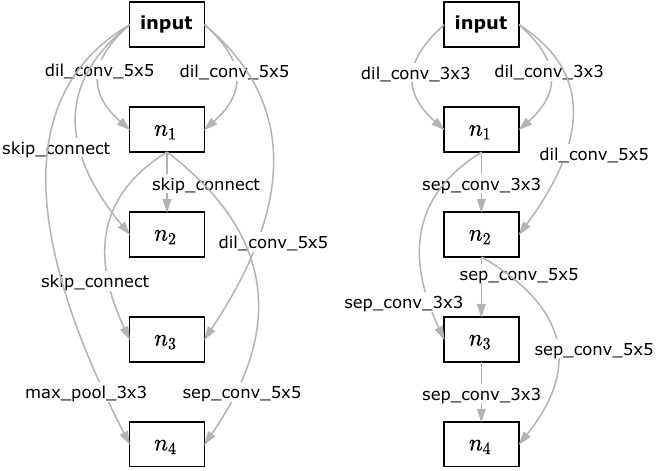}
    \caption{\textbf{Searched architectures of the normal cell (left) and reduction cell (right).} The searched cell contains four ordered nodes $\{n_1, n_2, n_3, n_4 \}$ and each node has two previous inputs. Each directed edge denotes the chosen operation. The output of the cell is the concatenation of the output of each node.}
    \label{fig:searched_architectures}
\end{figure}

\begin{figure}[t]
    \centering
    \includegraphics[width=\linewidth]{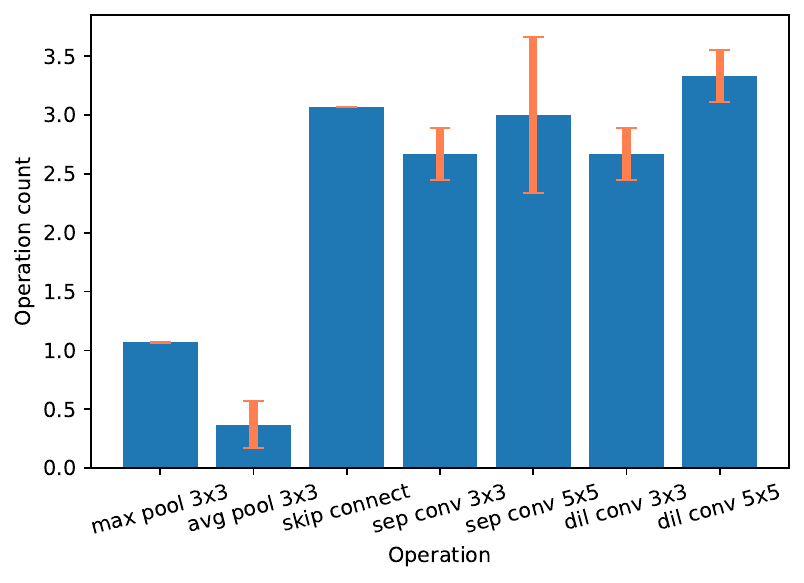}
    \caption{\textbf{Stability of searched architectures} with different initial random seeds.
    The operation percentage in each operation contains three bars, denoting the percentage of this operation in the searched architecture initialized by random seeds $0\sim2$ from left to right.}
    \label{fig:statistic_searched_architectures}
\end{figure}

\subsection{Search Space}
The design for the candidate operations is the same as DARTS~\cite{liu2018darts}, including average pooling with filter size $3 \times 3$, max pooling with filter size $3 \times 3$, separable convolutions with filter sizes $3 \times 3$, and $5 \times 5$, dilated separable convolutions with filter sizes $3 \times 3$, and $5 \times 5$, and skip connect. Note that our method is agnostic to the search space design and can be potentially extended to other search space designs. The search space mainly contains two types of cells---normal cells and reduction cells. Normal cells are foundational building blocks in our NAS framework. Reduction cells down-sample the input feature maps while normal cells maintain the size of feature maps after processing.

\subsection{Searched Architectures}
In this section, we visualize and analyze the searched architectures with the proposed method.


\paragraph{Patterns of searched architectures.}
The searched architecture is demonstrated in~\Cref{fig:searched_architectures}. 
As shown in~\Cref{fig:searched_architectures} (left), the searched normal cells tend to contain more large-kernel convolution layers and skip layers. This enlarges the receptive field of the searched architecture and encourages the network to learn hard-to-learn global feature representations instead of only easy-to-learn local features. 
On the other hand, the reduction cell is used for down-sampling the input feature map. The reduction cell searched by~\methodname~has  similar numbers of large and small kernel convolutional layers, as shown in~\Cref{fig:searched_architectures} (right). This enables the reduction cell to simultaneously learn local features and global semantic features, preventing it from generating high-dimensional representations solely from local or global features. This further validates the motivation of~\methodname~to improve generalization abilities.

\paragraph{Stability of searched architectures.}
We further test the stability of searched architectures against random initializations. The statistics of searched architectures with different random seeds are shown in ~\Cref{fig:statistic_searched_architectures}. As shown in the figure, the architectures searched by~\methodname~converge to the pattern that the percentages of convolutional layer and skip connect are higher, while the pooling layer is lower. This results demonstrate the stability of the pattern found by~\methodname~to improve OoD generalization ability.


\subsection{Visualization Results}
\begin{figure*}[t]
    \centering
    \includegraphics[width=\linewidth]{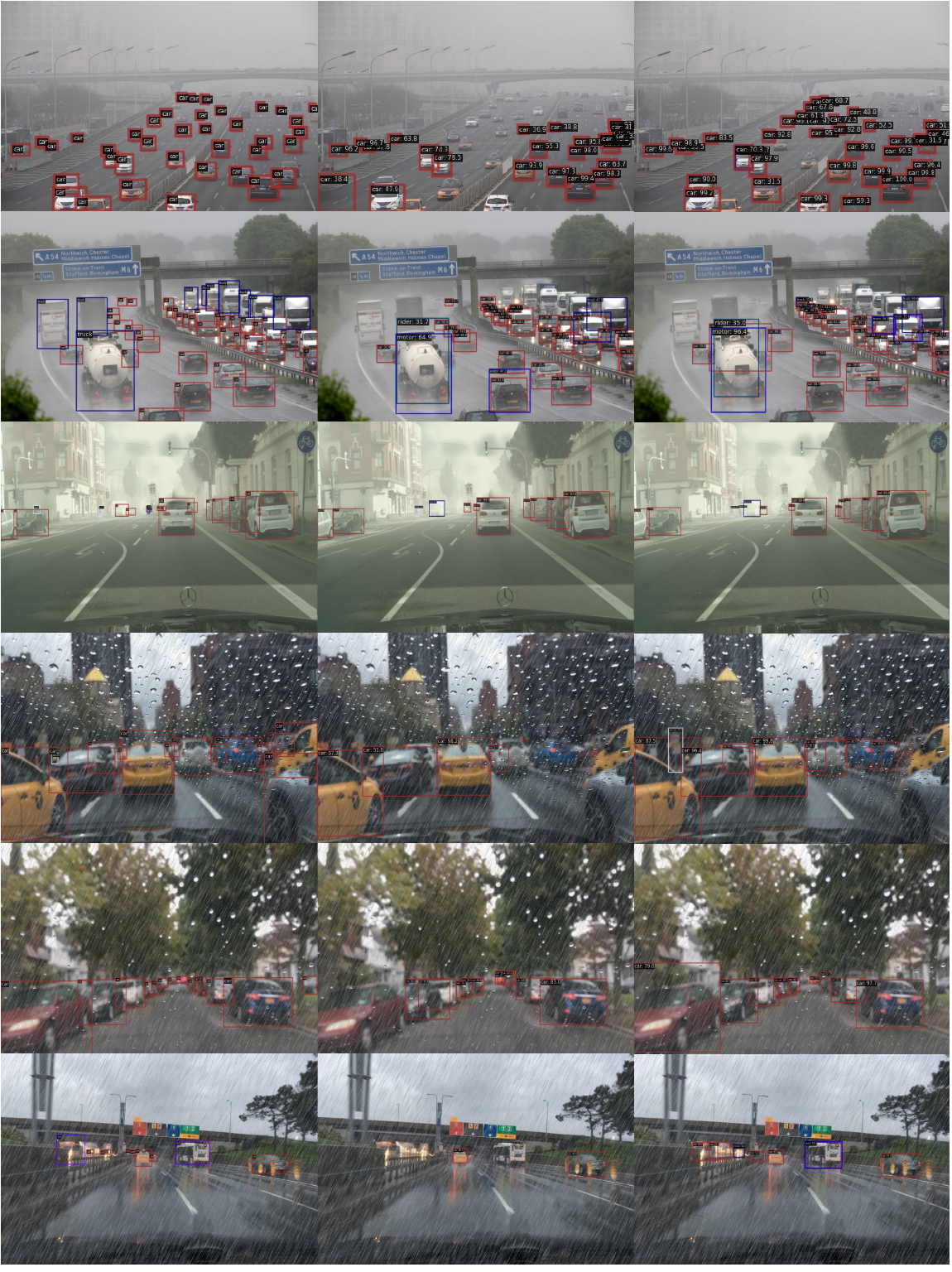}
    \caption{\textbf{Ground truth (left) and more inference results of Baselines (middle) and~\methodname~(right).} 
    The top three rows are on Daytime-Foggy and the bottom three rows are on Dusk-Rainy.
    The red boxes represent car objects, blue boxes represent truck objects, white boxes represent person objects and purple boxes represent bus objects.}
    \label{fig:more_infer_results_1}
\end{figure*}

\begin{figure*}[t]
    \centering
    \includegraphics[width=\linewidth]{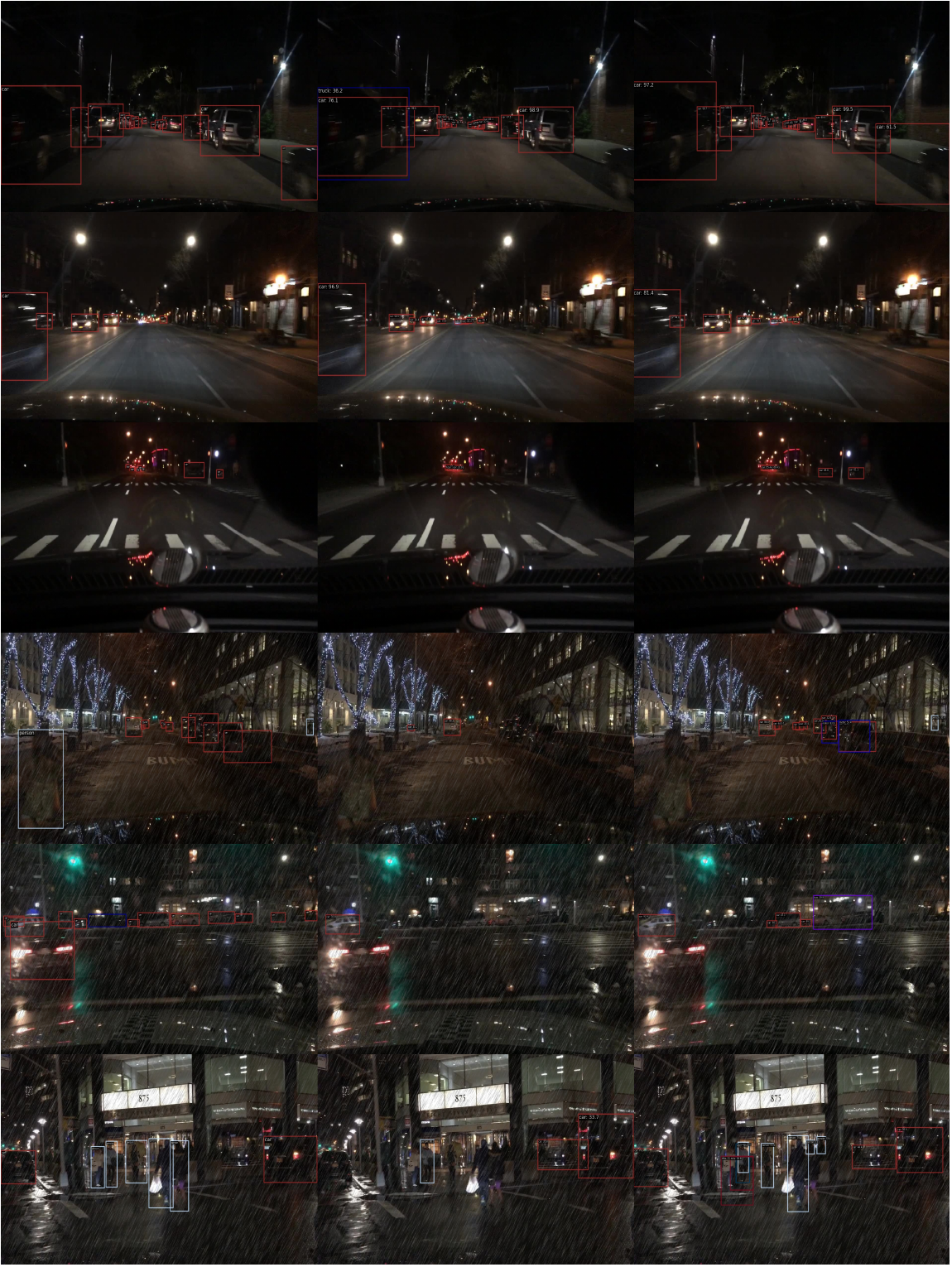}
    \caption{\textbf{Ground truth (left) and more inference results of Baselines (middle) and~\methodname~(right).}
    The top three rows are on Night-Sunny and the bottom three rows are on Night-Rainy.
    The red boxes represent car objects, blue boxes represent truck objects, and white boxes represent person objects.}
    \label{fig:more_infer_results_2}
\end{figure*}

We visualize examples of inference results in
\Cref{fig:more_infer_results_1} and~\Cref{fig:more_infer_results_2}. As shown in \Cref{fig:more_infer_results_1} and~\Cref{fig:more_infer_results_2}, the proposed method demonstrates robust object detection abilities under extremely challenging and unseen environments with only a single domain data for training.
For example, vehicles are hardly recognized if covered by fog, while our proposed~\methodname~accurately detect these vehicles compared with baselines, as shown in the first-row of~\Cref{fig:more_infer_results_1}.

\end{document}